%% file: CIRCLe.tex
\documentclass[runningheads]{llncs}
\usepackage{graphicx}

\usepackage{tikz}
\usepackage{comment}
\usepackage{amsmath,amssymb}
\usepackage{color}

\usepackage{cite}
\usepackage{booktabs}

\usepackage{arydshln}
\usepackage{multirow}
\usepackage[pagebackref]{hyperref}
\usepackage{gensymb}
\usepackage{amssymb}

\usepackage{pifont}
\newcommand{\cmark}{\ding{51}}
\newcommand{\xmark}{\ding{55}}

\DeclareMathOperator*{\argmaxC}{\arg\max}  
\usepackage{caption}
\usepackage{subcaption}

\usepackage{xspace}

\usepackage[accsupp]{axessibility}

\begin{document}
\pagestyle{headings}
\mainmatter

\newcommand{\cic}{CIRCLe\xspace}
\newcommand{\ourM}{Normalized Accuracy Range\xspace}
\newcommand{\ourMshort}{NAR\xspace}

\title{CIRCLe: Color Invariant Representation Learning for Unbiased Classification of Skin Lesions}

\titlerunning{Color Invariant Skin Lesion Representation Learning}
%
\author{Arezou Pakzad\inst{1} \and
Kumar Abhishek\inst{1} \and
Ghassan Hamarneh\inst{1}}

\authorrunning{A. Pakzad et al.}

\institute{School of Computing Science, Simon Fraser University, Canada\\
\email{\{arezou\_pakzad, kabhishe, hamarneh\}@sfu.ca}\\
}

\maketitle

\begin{abstract}

While deep learning based approaches have demonstrated expert-level performance in dermatological diagnosis tasks, they have also been shown to exhibit biases toward certain demographic attributes, particularly skin types (e.g., light versus dark), a fairness concern that must be addressed. We propose CIRCLe, a skin color invariant deep representation learning method for improving fairness in skin lesion classification. CIRCLe is trained to classify images by utilizing a regularization loss that encourages images with the same diagnosis but different skin types to have similar latent representations. Through extensive evaluation and ablation studies, we demonstrate CIRCLe's superior performance over the state-of-the-art when evaluated on 16k+ images spanning 6 Fitzpatrick skin types and 114 diseases, using classification accuracy, equal opportunity difference (for light versus dark groups), and normalized accuracy range, a new measure we propose to assess fairness on multiple skin type groups. Our code is available at \url{https://github.com/arezou-pakzad/CIRCLe}.

\keywords{
Fair AI  \and Skin Type Bias \and Dermatology \and Classification \and Representation Learning.
}
\end{abstract}

\section{Introduction}

Owing to the advancements in deep learning (DL)-based data-driven learning paradigm, convolutional neural networks (CNNs) can be helpful decision support tools in healthcare. This is particularly true for dermatological applications where recent research has shown that DL-based models can reach the dermatologist-level classification accuracies for skin diseases~\cite{esteva2017dermatologist,haenssle2018,brinker2019} while doing so in a clinically interpretable manner~\cite{barata2019deep,liu2020deep}.
However, this data-driven learning paradigm that allows models to automatically learn meaningful representations from data leads DL models to mimic biases found in the data, i.e., biases in the data can propagate through the learning process and result in an inherently biased model, and consequently in a biased output.

Most public skin disease image datasets are acquired from demographics consisting primarily of fair-skinned people. However, skin conditions exhibit vast visual differences in manifestations across different skin types~\cite{brownskinmatters}. Lighter skinned populations suffer from over-diagnosis of melanoma~\cite{adamson2022estimating} while darker skinned patients get diagnosed at later stages, leading to increased morbidity and mortality~\cite{agba2014cancer}. Despite this, darker skin is under-represented in most publicly available data sets~\cite{lester2020absence,kinyanjui2020fairness}, reported studies~\cite{daneshjou2021lack}, and in dermatology textbooks~\cite{adelekun2021skin}.
Kinyanjui et al.~\cite{kinyanjui2020fairness} performed an analysis on two popular benchmark dermatology datasets: ISIC 2018 Challenge dataset~\cite{codella2019skin} and SD-198 dataset~\cite{sun2016benchmark}, to understand the skin type representations.
They measured the individual typology angle (ITA), which measures the constitutive pigementation of skin images~\cite{osto2022individual}, to estimate the skin tone on these datasets, and  found that the majority of the images in the two datasets ITA values between $34.8\degree$ and $48\degree$, which are associated with lighter skin. This is consistent with the under-representation of darker skinned populations in these datasets.
It has been shown that CNNs perform best at classifying skin conditions for skin types that are similar to those they were trained on~\cite{groh2021evaluating}. Thus, the data imbalance across different skin types in the majority of the skin disease image datasets can manifest as racial biases in the DL models' predictions, leading to racial disparities~\cite{adamson2018machine}. However, despite these well-documented concerns, very little research has been directed towards evaluating these DL-based skin disease diagnosis models on diverse skin types, and therefore, their utility and reliability as disease screening tools remains untested.

Although research into algorithmic bias and fairness has been an active area of research, interest in fairness of machine learning algorithms in particular is fairly recent. Multiple studies have shown the inherent racial disparities in machine learning algorithms' decisions for a wide range of areas: pre-trial bail decisions~\cite{kleinberg2018human}, recidivism~\cite{angwin2016machine}, healthcare~\cite{obermeyer2019}, facial recognition~\cite{buolamwini2018gender}, and college admissions~\cite{kleinberg2018algorithmic}. Specific to healthcare applications, previous research has shown the effect of dataset biases on DL models' performance across genders and racial groups in cardiac MR imaging~\cite{esther2021}, chest X-rays~\cite{larrazabal2020gender,seyyed2020chexclusion,seyyed2021underdiagnosis}, and skin disease imaging~\cite{groh2021evaluating}. Recently, Groh et al.~\cite{groh2021evaluating} showed that CNNs are the most accurate when classifying skin diseases manifesting on skin types similar to those they were trained on.

Learning domain invariant representations, a predominant approach in domain generalization~\cite{muandet2013}, attempts to learn data distributions that are independent of the underlying domains, and therefore addresses the issue of training models on data from a set of source domains that can generalize well to previously unseen test domains. Domain invariant representation learning has been used in medical imaging for histopathology image analysis~\cite{lafarge2019learning} and for learning domain-invariant shape priors in segmentation of prostrate MR and retinal fundus images~\cite{liu2022single}. On the other hand, previous works on fair classification and diagnosis of skin diseases have relied on skin type detection and debiasing~\cite{bevan2022detecting} and classification model pruning~\cite{wu2022fairprune}.

One of the common definitions of algorithmic fairness for classification tasks, based on measuring statistical parity, aims to seek independence between the bias attribute (also known as the protected attribute; i.e., the skin type for our task) and the model's prediction (i.e., the skin disease prediction). Our proposed approach, \textbf{C}olor \textbf{I}nvariant \textbf{R}epresentation learning for unbiased \textbf{C}lassification of skin \textbf{Le}sions (\textbf{\cic}), employs a color-invariant model that is trained to classify skin conditions independent of the underlying skin type. 
In this work, we aim to mitigate the skin type bias learnt by the CNNs and reduce the accuracy disparities across skin types.
We address this problem by enforcing the feature representation to be invariant across different skin types.
We adopt a domain-invariant representation learning method~\cite{nguyen2021domain} and modify it to transform skin types from clinical skin images 
and propose a color-invariant skin condition classifier. 
In particular, we make the following contributions:
\begin{itemize}
    \item To the best of our knowledge, this is the first work that uses skin type transformations and skin color-invariant disease classification to tackle the problem of skin type bias present in large scale clinical image datasets and how these biases permeate through the prediction models.
    \item We present a new state-of-the-art classification accuracy over 114 skin conditions and 6 Fitzpatrick skin types (FSTs) from the Fitzpatrick17K dataset. While previous works had either limited their analysis to a subset of diagnoses~\cite{bevan2022detecting} or less granular FST labels~\cite{wu2022fairprune}, our proposed method achieves superior performance over a much larger set of diagnoses spanning over all the FST labels.
    \item We provide a comprehensive evaluation of our proposed method, \cic, on 6 different CNN architectures, along with ablation studies to demonstrate the efficacy of the proposed domain regularization loss. Furthermore, we also assess the impact of varying the size and the FST distribution of the training dataset partitions on the generalization performance of the classification models.
    \item Finally, we propose a new fairness metric called \ourM that, unlike several existing fairness metrics, works with multiple protected groups (6 different FSTs in our problem).
\end{itemize}

\section{Method}

\begin{figure}[!ht]
\centering
     \includegraphics[width=1.0\textwidth]{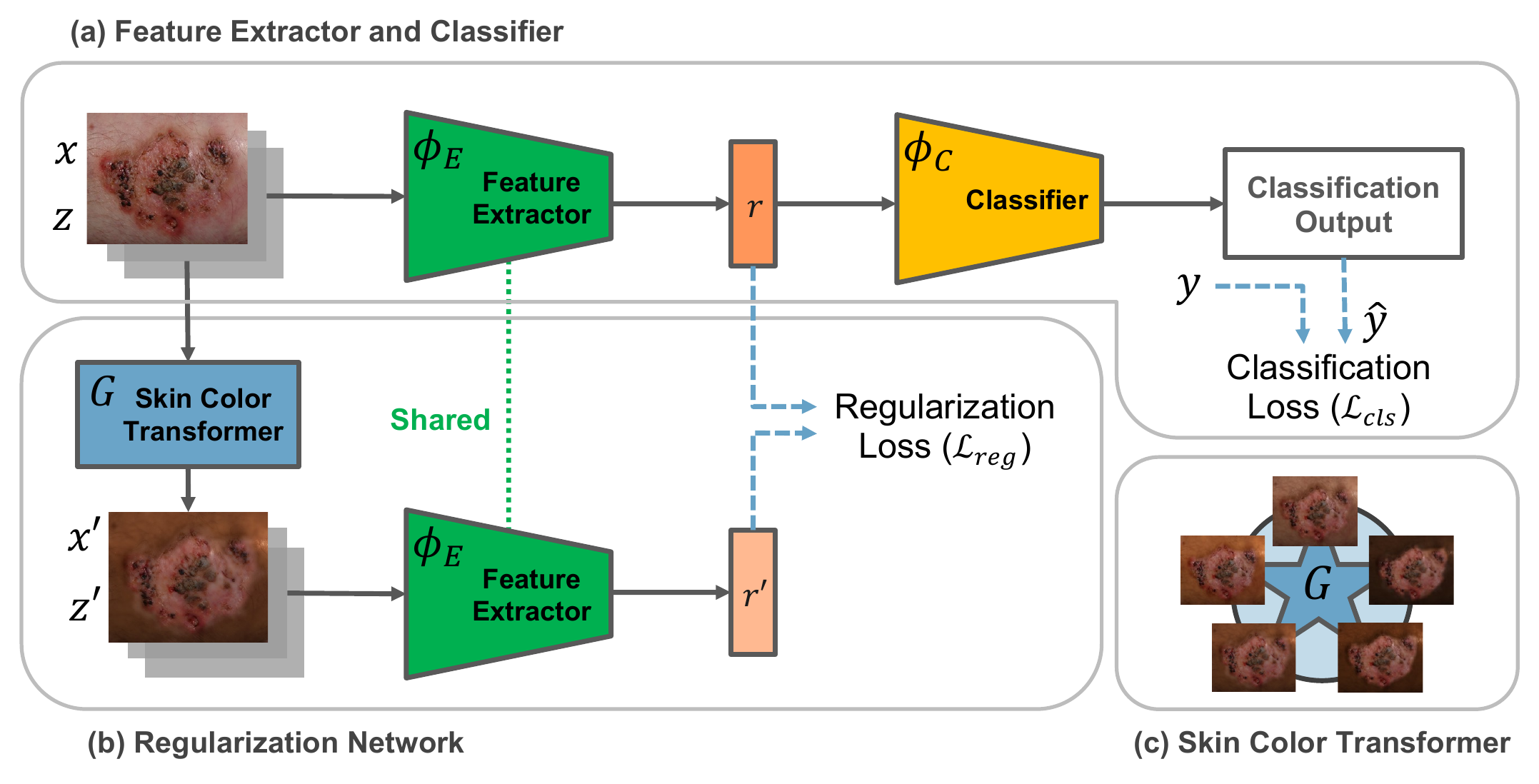}
      \caption{
      Overview of \cic.
      (a) The skin lesion image $x$ with skin type $z$ and diagnosis label $y$ is passed through the feature extractor $\phi_E$. The learned representation $r$ goes through the classifier $\phi_C$ to obtain the predicted label $\hat{y}$.
      The classification loss enforces the correct classification objective.
      (b) The skin color transformer ($G$), transforms $x$ with skin type $z$ into $x'$ with the new skin type $z'$. The generated image $x'$ is fed into the feature extractor to get the representation $r'$. 
      The regularization loss enforces $r$ and $r'$ to be similar.
      (c) The skin color transformer's schematic view with the possible transformed images, where one of the possible transformations is randomly chosen for generating $x'$.
      }
      \label{fig:model}
\end{figure}

\subsection{Problem Definition}
\label{sec:formal_def}
Given a dataset $\mathcal{D} = \{X, Y, Z\}$, consider $x_i, y_i, z_i$ to be the input, the label, and the protected attribute for the $i$\textsuperscript{th} sample respectively, where we have $M$ classes ($|Y| = M$) and $N$ protected groups ($|Z| = N$). Let $\hat{y_i}$ denote the predicted label of sample $i$. Our goal is to train a classification model $f_{\theta}(\cdot)$ parametrized by $\theta$ that maps the input $x_i$ to the final prediction $\hat{y_i} = f_{\theta}(x_i)$, such that (1) the prediction $\hat{y_i}$ is \emph{invariant} to the protected attribute $z_i$ and (2) the model's classification loss is minimized.

\subsection{Feature Extractor and Classifier}
\label{sec:E_and_C}
In the representation learning framework, the prediction function $\hat{y_i} = f_{\theta}(x_i)$ is obtained as a composition $\hat{y_i} = \phi_C \circ \phi_E (x_i)$ of a feature extractor $r_i = \phi_E (x_i)$, where $r_i \in \mathbb{R}^p$ is a learned representation of data $x_i$, and a classifier $\hat{y_i} = \phi_C (r_i)$, predicting the label $\hat{y_i}$, given the representation $r_i$ (Figure~\ref{fig:model}(a)). Thus, we aim to learn a feature representation $r$ that is invariant to the protected attributes, and hypothesize that this will lead to better generalization for classification.

\subsection{Regularization Network}
\label{sec:regnet}
Inspired by the method proposed by Nguyen et al.~\cite{nguyen2021domain}, we use a generative modelling framework to learn a function $g$ that transforms the data distributions between skin types.
To this end, we employ a method to synthesize a new image corresponding to a given input image with the subject's skin type in that image changed according to the desired Fitzpatrick skin type (FST) score. 
We call this model our Skin Color Transformer.
After training the Skin Color Transformer model, we introduce an auxiliary loss term to our learning objective, whose aim is to enforce the domain invariance constraint. (Figure~\ref{fig:model}(b))

\subsubsection{Skin Color Transformer.} 
We learn the function $G$ that performs image-to-image transformations between skin type domains.
To this end, we use a Star Generative Adversarial Network (StarGAN)~\cite{choi2018stargan}.
The goal of the StarGAN is to learn a unified network $G$ (generator) that transforms the data density among multiple domains.
In particular, the network $G(x, z, z')$ 
transforms an image $x$ from skin type $z$ to skin type $z'$.
The generator's goal is to fool the discriminator $D$ into classifying the transformed image as the destination skin type $z'$.
In other words, the equilibrium state of StarGAN
is when $G$ successfully transforms the data density of the original skin type to that of the destination skin type.
After training, we use $G(., z, z')$ as the Skin Color Transformer.
This model takes the image $x_i$ with skin type $z_i$ as the input, along with a target skin type $z_j$ and synthesizes a new image $z_i' = G(x_i, z_i, z_j)$ similar to $x_i$, only with the skin type of the image changed in accordance with $z_j$.

\subsubsection{Domain Regularization Loss.}
In the training process of the disease classifier, for each input image $x_i$ with skin type $s_i$, we randomly select another skin type $s_j \neq s_i$, and use the Skin Type Transformer to synthesize a new image $x_i' = G(x_i, s_i, s_j)$.
After that, we obtain the latent representations $r_i = \phi_E (x_i)$, and $r_i' = \phi_E (x_i')$ for the original image and the synthetic image respectively. 
Then we enforce the model to learn similar representations for $r_i$ and $r_i'$ by adding a regularization loss term to the overall loss function of the model:
\begin{equation}
    \mathcal{L}_{total} = \mathcal{L}_{cls} + \lambda \mathcal{L}_{reg}    
\end{equation}
where $\mathcal{L}_{cls}$ is the prediction loss of the network that predicts $\hat{y_i}$ given $r_i = \phi_E (x_i)$, and $\mathcal{L}_{reg}$ is the regularization loss. In this equation, $\lambda \in [0, 1]$ is a hyper-parameter controlling the trade-off between the classification and regularization losses. We define $\mathcal{L}_{reg}$ as the distance between the two representations $r_i$ and $r_i'$ to enforce the invariant condition. 
In our implementation, we use cross entropy as the classification loss $\mathcal{L}_{cls}$:
\begin{equation}
    \mathcal{L}_{cls} = -\sum_{j = 1}^{M} y_{ij} \ \mathrm{log} (\hat{y}_{ij}),
\end{equation}
\noindent where $y_{ij}$ is a binary indicator (0 or 1) if class label $j$ is the correct classification for the sample $i$ and $\hat{y}_{ij}$ is the predicted probability the sample $i$ is of class $j$. The final predicted class $\hat{y}_{i}$ is calculated as
\begin{equation}
    \hat{y}_{i} = \argmaxC_{j} \ \hat{y}_{ij}.
\end{equation}
We use squared error distance for computing the regularization loss $\mathcal{L}_{reg}$:
\begin{equation}
    \mathcal{L}_{reg} = || r_i - r_i' ||_2^2.
\end{equation}

\section{Experiments}
\subsection{Dataset}
\begin{figure}[!ht]
\centering
     \includegraphics[width=1.0\textwidth]{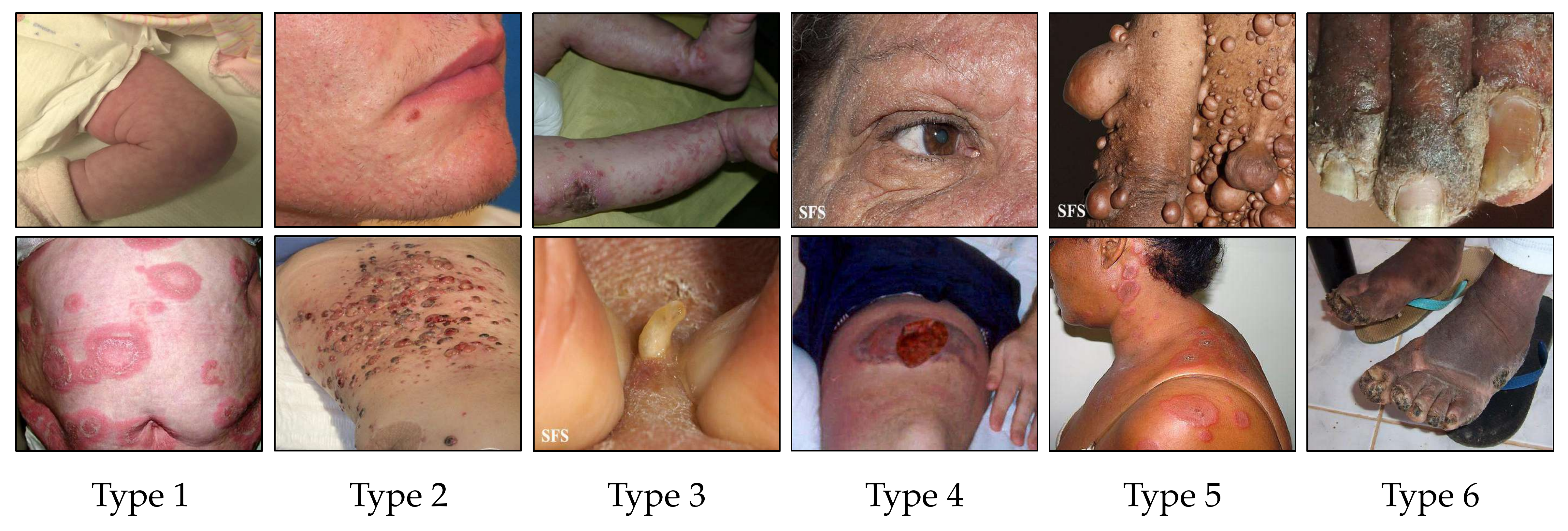}
      \caption{
      Sample images of all six Fitzpatrick skin types (FSTs) from the Fitzpatrick17K dataset~\cite{groh2021evaluating}. Notice the wide varieties in disease appearance, field of view, illumination, presence of imaging artifacts including non-standard background consistent with clinical images in the wild, and watermarks on some images.
      }
      \label{fig:FST_overview}
\end{figure}

\begin{figure}[ht!]
     \centering
     \begin{subfigure}[b]{0.58\textwidth}
         \centering
         \includegraphics[width=\textwidth]{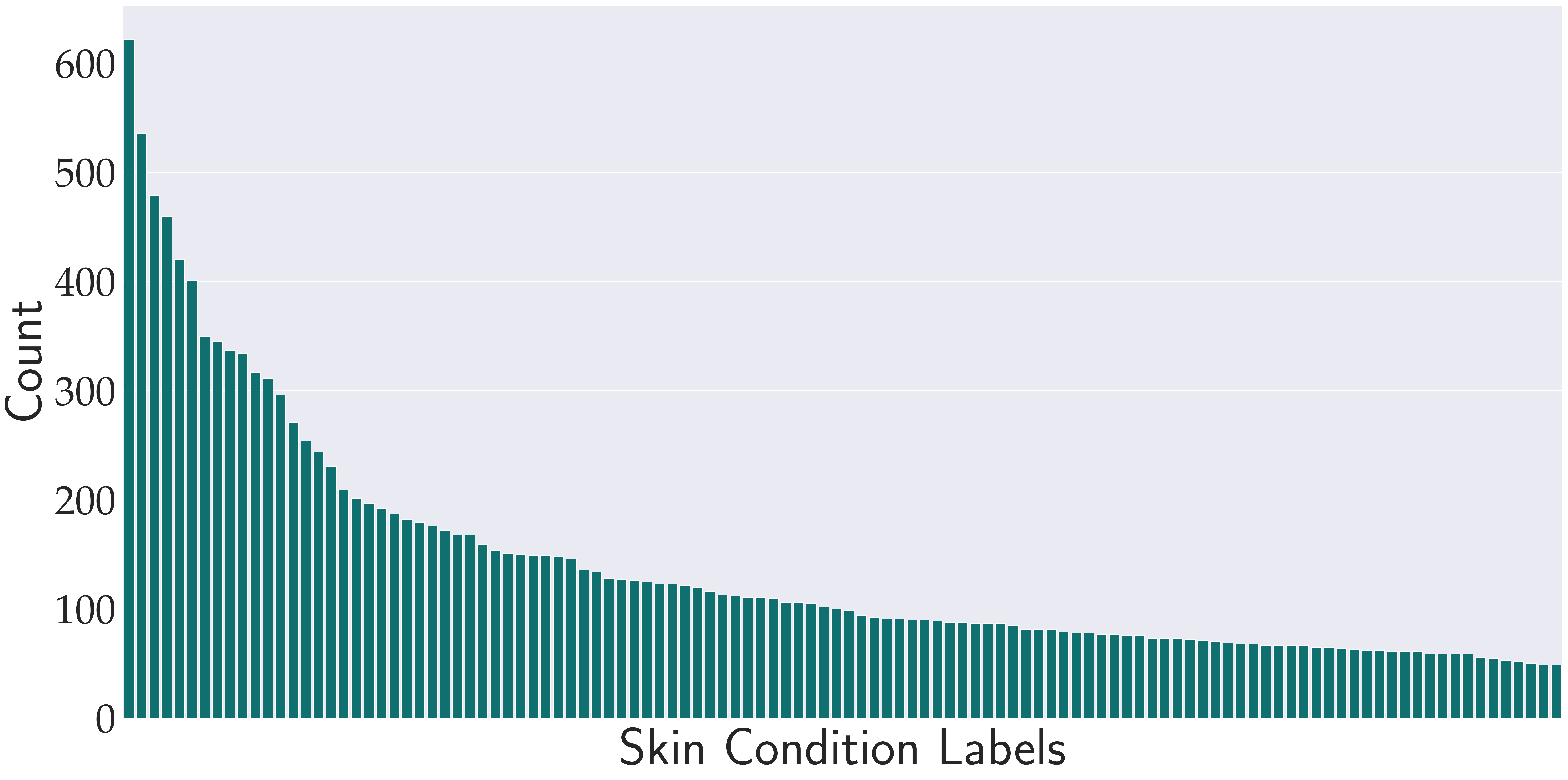}
         \caption{}
     \end{subfigure}
     \hfill
     \begin{subfigure}[b]{0.4\textwidth}
         \centering
         \includegraphics[width=\textwidth]{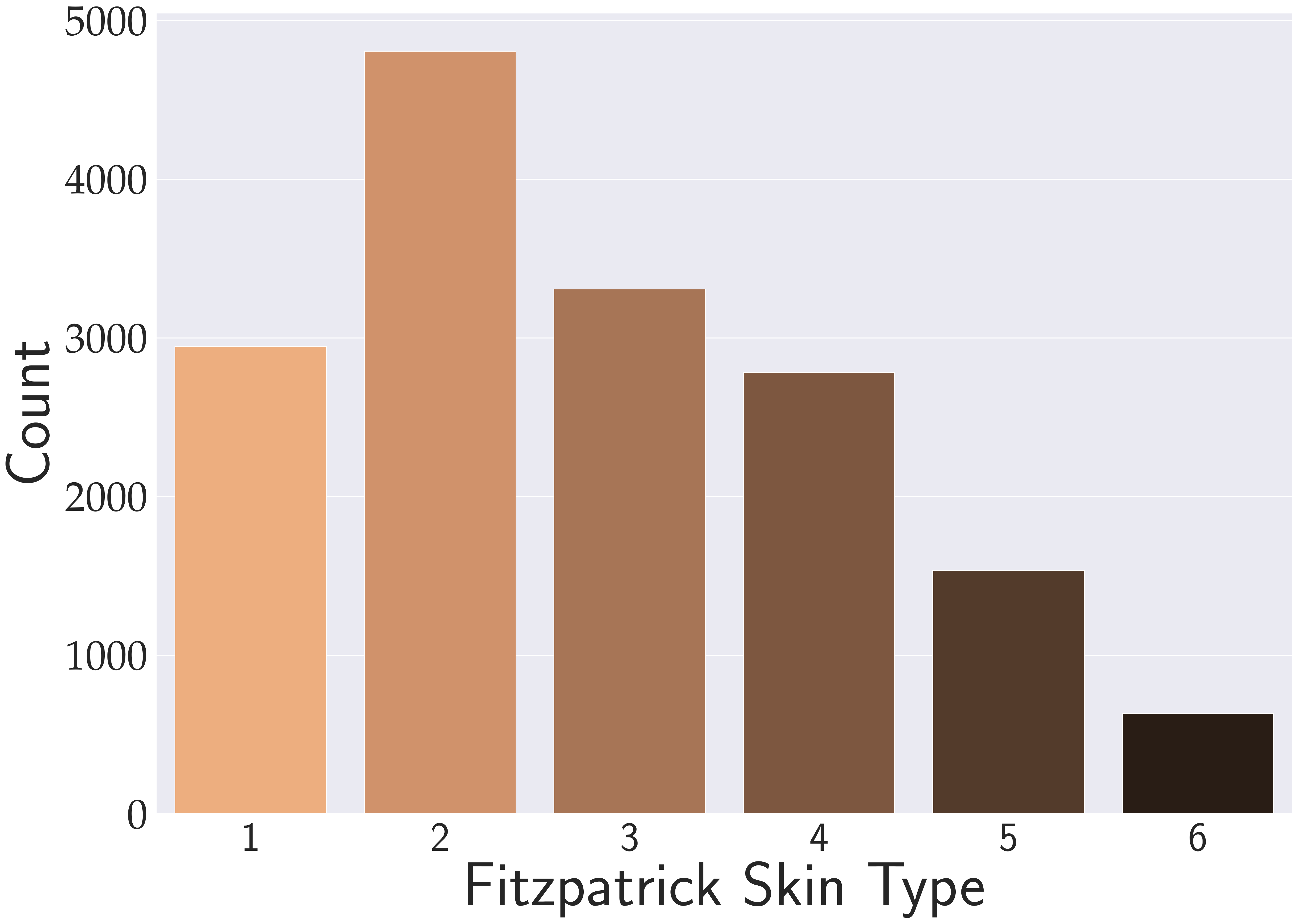}
         \caption{}
     \end{subfigure}
        \caption{Visualizing the distribution of (a) the skin condition labels and (b) the Fitzpatrick skin type (FST) labels in the Fitzpatrick17K dataset. Notice that the number of images across different skin conditions is not uniformly distributed. Moreover, the number of images is considerably lower for darker skin types.}
        \label{fig:distribution}
\end{figure}
We evaluate the performance of the proposed method on the Fitzpatrick17K dataset~\cite{groh2021evaluating}.
The Fitzpatrick17K dataset contains 16,577 clinical images with skin condition labels and skin type labels based on the Fitzpatrick scoring system~\cite{archderm1988}.
The dataset includes 114 conditions with at least 53 images (and a maximum of 653 images) per skin condition, as shown in Figure~\ref{fig:distribution} (a). 
The images in this dataset are annotated with (FST) labels by a team of non-dermatologist annotators.
Figure~\ref{fig:FST_overview} shows some sample images from this dataset along with their skin types.
The Fitzpatrick labeling system is a six-point scale originally developed for classifying sun reactivity of skin and adjusting clinical medicine according to skin phenotype~\cite{archderm1988}. In this scale, the skin types are categorized in six levels from 1 to 6, from lightest to darkest skin types.
Although Fitzpatrick labels are commonly used for categorizing skin types, we note that not all skin types are represented by the Fitzpatrick scale.~\cite{ware2020racial}.\par
In the Fitzpatrick17K dataset, there are significantly more images of light skin types than dark skin.
There are 11,060 images of \textit{light} skin types (FSTs 1, 2, and 3), and 4,949 images of \textit{dark} skin types (FSTs 4, 5, and 6), as shown in Figure~\ref{fig:distribution} (b).

\subsection{Implementation Details}

\subsubsection{Dataset Construction Details.}
We randomly select $70\%$, $10\%$, and $20\%$ of the images for the train, validation, and test splits, where the random selection is stratified on skin conditions.
We repeat the experiments with five different random seeds for splitting the data.
A series of transformations are applied to the training images which include: resize to $128 \times 128$ resolution, random rotations in [$-15\degree$, $15\degree$], and random horizontal flips. 
We also use ImageNet~\cite{deng2009imagenet} training partition's mean and standard deviation values to normalize our images for training and evaluation.

\subsubsection{Feature Extractor and Classifier's Details.}
We choose VGG-16~\cite{simonyan2014very} pre-trained on ImageNet as our base network.
We use the convolutional layers of VGG-16 as the feature extractor $\phi_E$.
We replace the VGG-16's fully-connected layers with a fully connected 256-to-114 layer as the classifier $\phi_C$.
We train the network for 100 epochs with plain stochastic gradient descent (SGD) using learning rate 1e-3, momentum 0.9, minibatch size 16, and weight decay 1e-3.
We report the results for the epoch with the highest accuracy on the validation set.

\subsubsection{StarGAN Details.}
StarGAN~\cite{choi2018stargan} implementation is taken from the authors' original source code with no significant modifications.
We train StarGAN on the same train split used for training the classifier.
As for the training configurations we use a minibatch size of 16.
We train the StarGAN for 200,000 iterations and use the Adam~\cite{kingma2014adam} optimizer with a learning rate of 1e-4.
For training the StarGAN's discriminator, we use cross entropy loss.

\subsubsection{Model Training and Evaluation Setup.}
We use the PyTorch library~\cite{pytorch2019} to implement our framework and train all our models on a workstation with AMD Ryzen 9 5950X processor, 32 GB of memory, and Nvidia GeForce RTX 3090 GPU with $24$ GB of memory.

\subsection{Metrics}    \label{subsec:metrics}
We aim for an \textit{accurate} and \textit{fair} skin condition classifier.
Therefore, we assess our method's performance using metrics for both accuracy and fairness.
We use the well-known and commonly-used recall, F1-score, and accuracy metrics for evaluating our model's classification performance. 
For fairness, we use the equal opportunity difference (EOD) metric~\cite{Hardt2016EOD}.
EOD measures the difference in true positive rates (TPR) for the two protected groups.
Let $TPR_{z}$ denote true positive rate of group $z$ and $z \in \{0, 1\}$.
Then $EOD$ can be computed as:

\begin{equation}
    EOD = |TPR_{z=0} - TPR_{z=1}|.
\end{equation}
A value of $0$ implies both protected groups have equal benefit.
Given that the above metric (and other common fairness metrics in the literature~\cite{Hardt2016EOD, dwork2012fairness, bellamy2019ai}) are defined for two groups: privileged and under-privileged, w.r.t the protected attribute, we adopt the light (FSTs 1, 2, and 3) versus dark (FSTs 4, 5, and 6) as the two groups.

Additionally, to measure fairness in the model's accuracy for multiple groups of skin types, we assess the accuracy (ACC) disparities across all the six skin types by proposing the \ourM (\ourMshort) as follows:

\begin{equation}
    \ourMshort = \frac{ACC_{max} - ACC_{min}}{mean(ACC)}, \\
\end{equation}
where $ACC_{max}$ and $ACC_{min}$ are the maximum and minimum accuracy achieved across skin types and $mean(ACC)$ is the mean accuracy across skin types, i.e.:
\begin{equation}
\begin{gathered}
    ACC_{max} = max\{ACC_i: 1 \leq i \leq N\}, \\
    ACC_{min} = min\{ACC_i: 1 \leq i \leq N\}, \\
    mean(ACC) = \frac{1}{N} \sum_{i=1}^{N} ACC_i
  \end{gathered}
\end{equation}
A perfectly fair performance of a model would result in equal accuracy across the different protected groups on a test set, i.e. $ACC_{max} = ACC_{min}$, leading to $NAR = 0$.
As the accuracies across protected groups diverge, $ACC_{max} > ACC_{min}$, \ourMshort will change even if the mean accuracy remains the same, thus indicating that the model's fairness is also changed.
Moreover, \ourMshort also takes into account the overall mean accuracy: this implies that in cases where the accuracies range $(ACC_{max} - ACC_{min})$ is the same, the model with the overall higher accuracy leads to a lower \ourMshort, which is desirable.
In our quantitative results, we report EOD for completeness; however, it is not an ideal measure, given it is restricted to only two protected groups whereas we have six. Therefore, we focus our attention on \ourMshort.

\subsection{Models}
\subsubsection{Baseline.}
For evaluating our method, we compare our results with the method proposed by Groh et al.~\cite{groh2021evaluating}, which has the current state-of-the-art performance on the Fitzpatrick17K dataset.
We call their method the \textit{Baseline}.
To obtain a fair comparison, we use the same train and test sets they used.  

\input{tables/table5}

\subsubsection{Improved Baseline (Ours).}
In order to evaluate the effectiveness of the color-invariant representation learning process, we perform an ablation study, in which we remove the regularization loss $\mathcal{L}_{reg}$ from the learning objective of the model and train the classifier with only the classification objective. 
We call this model the \textit{Improved Baseline}.

\subsubsection{\cic (Ours).}
The proposed model for unbiased skin condition classification, \cic, is composed of two main components: the feature extractor and classifier, and the regularization network (Fig. \ref{fig:model}).

\subsubsection{Multiple Backbones.}

To demonstrate the efficacy of our method, we present evaluation with several other backbone architectures in addition to VGG-16~\cite{simonyan2014very} used by Groh et al.~\cite{groh2021evaluating}. In particular, we use MobileNetV2~\cite{sandler2018mobilenetv2}, MobileNetV3-Large (referred to as MobileNetV3L hereafter)~\cite{howard2019searching}, DenseNet-121~\cite{huang2017densely}, ResNet-18~\cite{he2016deep}, and ResNet-50~\cite{he2016deep}, thus covering a wide range of CNN architecture families and a considerable variety in model capacities, i.e. from 2.55 million parameters in MobileNetV2 to 135.31 million parameters in VGG-16 (Table~\ref{table:backboneparams}).

For all the models, we perform an ablation study to evaluate if adding the regularization loss $\mathcal{L}_{reg}$ helps improve the performance.

\section{Results and Analysis}
\subsection{Classification and Fairness Performance.}
\label{sec:performance_results}
\input{tables/table1}
Table \ref{table:result} shows the accuracy and fairness results for the proposed method in comparison with the baseline.
From the table, we can see that our Improved Baseline method recognizably outperforms the baseline method in accuracy and fairness.
By using a powerful backbone and a better and longer training process, we more than doubled the classification accuracy on the Fitzpatrick17K dataset for all the skin types.
This indicates that the choice of the base classifier and training settings plays a significant role in achieving higher accuracy rates on the Fitzpatrick17K dataset.
Moreover, we can see that \cic further improves the performance of our Improved Baseline across all the skin types, as well as the overall accuracy.
This significant improvement demonstrates the effectiveness of the color-invariant representation learning method in increasing the model's generalizability.
This observation shows that when the model is constrained to learn similar representations from different skin types that the skin condition appears on, it can learn richer features from the disease information in the image, and its overall performance improves.
In addition, \cic shows improved fairness scores (lower EOD and lower \ourMshort), which indicates that the model is less biased. To the best of our knowledge, we set a new state-of-the-art performance on the Fitzpatrick17K dataset for the task of classifying the 114 skin conditions.

Different model architectures may show different disparities across protected groups~\cite{prince2019tutorial}.

We can see in Table~\ref{table:backbones} that the color-invariant representation learning (i.e. with the regularization loss $\mathcal{L}_{reg}$ activated) significantly improves the accuracy and fairness results in different model architecture choices across skin types, which indicates the effectiveness of the proposed method independently from the backbone choice and its capacity. We can see that while the regularization loss does not necessarily improve the EOD for all the backbones, EOD is not the ideal measure of fairness for our task since as explained in Section~\ref{subsec:metrics}, it can only be applied to a lighter-versus-darker skin tone fairness assessment. However, employing the regularization loss does improve the \ourMshort for all the backbone architectures.
\input{tables/table2}
\input{tables/table3}
\subsection{Domain Adaptation Performance}

For evaluating the model's performance on adapting to unseen domains, we perform a ``two-to-other'' experiment, where we train the model on all the images from two FST domains and test it on all the other FST domains.
Table~\ref{table:2skin} shows the performance of our model for this experiment.
\cic recognizably improves the domain adaptation performance in comparison with the Baseline and Improved Baseline, demonstrating the effectiveness of the proposed method in learning a color-invariant representation.

\subsection{Classification Performance Relation with Training Size}
\label{sec:partial}
\begin{figure}[h!]
\centering
     \includegraphics[width=1.0\textwidth]{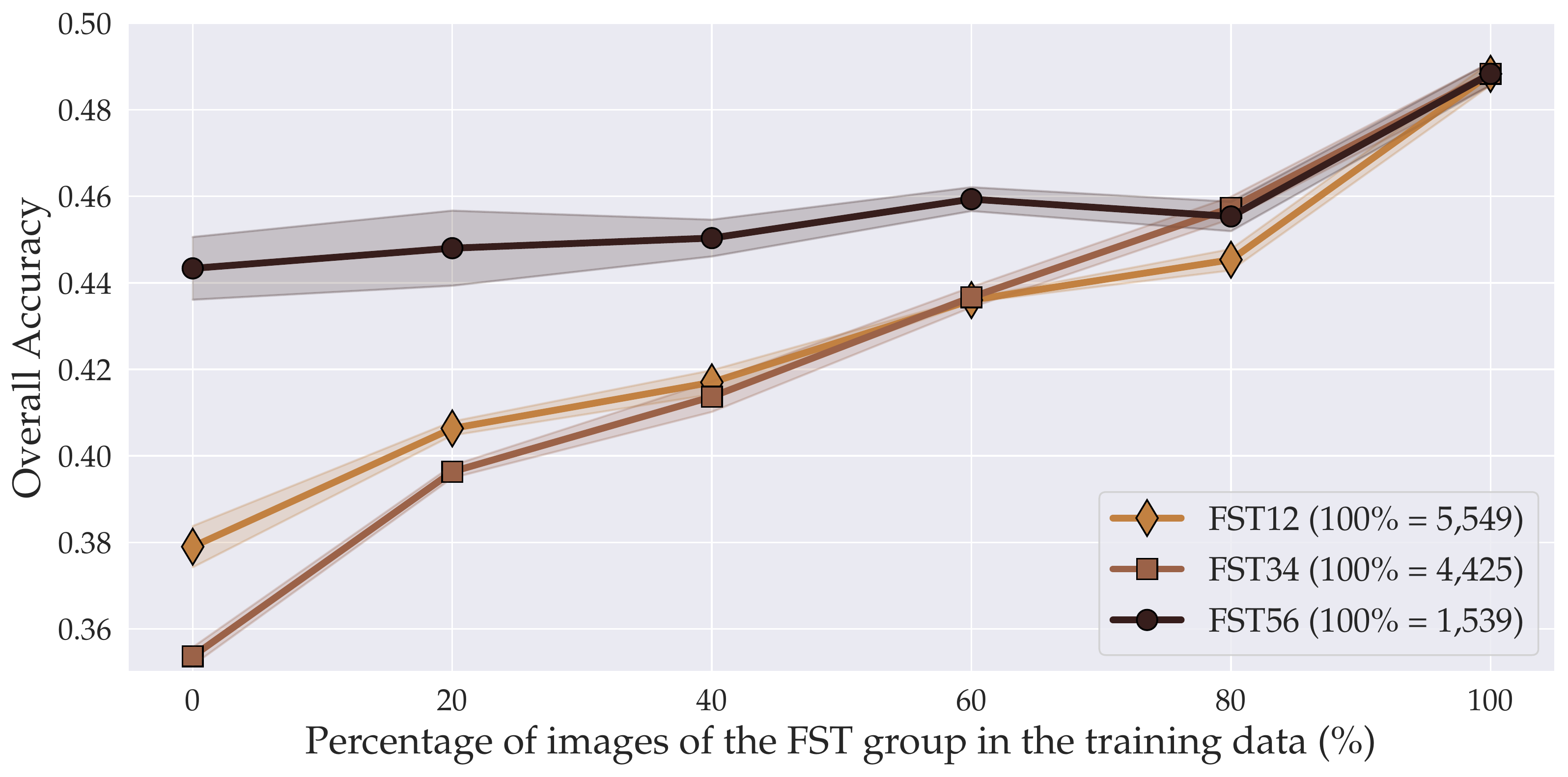}
      \caption{
      Classification performance of \cic on the test set as the number of training images of the FST groups increases.
      Each FST group line plot indicates the series of experiments in which the percentage of number of training images of that FST group changes as the rest of the training images remain idle.
      The rightmost point in the plot, with $100\%$, is identical for all the FST groups, which is the overall accuracy achieved by \cic in Table \ref{table:result}.
      The std. dev. error band, illustrated in the figure, is computed by repetition of experiments with three different random seeds. 
      }
      \label{fig:partial}
\end{figure}

\input{tables/table4}

As \cic's performance improvement and effectiveness in comparison with the baselines is established in Section~\ref{sec:performance_results}, we further analyze the relation of \cic's classification performance with the percentage of images of the FST groups in the training data.
To this end, we consider the FST groups of light skin types (FSTs 1 and 2) with 5,549 images, medium skin types (FSTs 3 and 4)  with 4,425 images, and dark skin types (FSTs 5 and 6) with  1,539 images in the training set.
For each FST group, we gradually increase the number of images of that group in the training set, while the number of training images in other groups remains unchanged, and report the model's overall accuracy on the test set.
The total number of training images for each of these experiments is provided in Table~\ref{table:partial}.
As we can see in Figure~\ref{fig:partial}, as the number of training images in a certain FST group increases, the overall performance improves, which is expected since DL-based models generalize better with larger training datasets.
However, we can see that for the least populated FST group, i.e., dark skin types (FST56) with $13\%$ of the training data, our method demonstrates a more robust performance across experiments, and even with $0\%$ training data of FST56, it achieves a relatively high classification accuracy of $0.443$.
In addition, note that in these experiments, FST groups with lower number of images in the dataset, would have a larger number of total training images, since removing a percentage of them from the training images will leave a larger portion of images available for training (see Table~\ref{table:partial}).
This indicates that when the number of training images is large enough, even if images of a certain skin type are not available, or are very limited, our model can perform well overall.
This observation signifies our method's ability to effectively utilize the disease-related features in the images from the training set, independently from their skin types, as well as the ability to generalize well to minority groups in the training set.

\section{Discussion and Future Work}
\begin{figure}[ht]
\centering
     \includegraphics[width=1\textwidth]{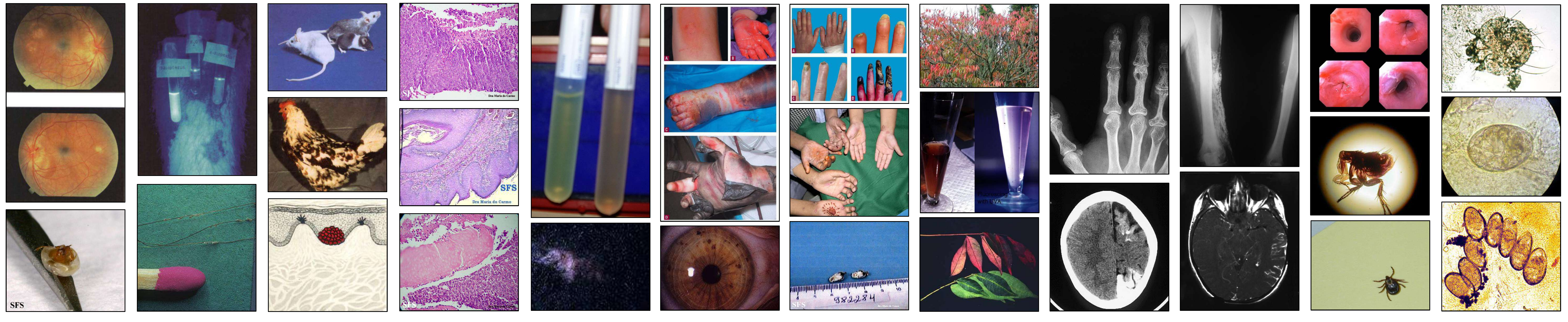}
      \caption{Sample erroneous images from the Fitzpatrick17K dataset that are not clinical images of skin conditions, but are included in the dataset and are wrongly labeled with skin conditions.}
      \label{fig:noise}
\end{figure}

In order to develop fair and accurate DL-based data-driven diagnosis methods in demotology, we need annotated datasets that include a diversity of skin types and a range of skin conditions.
However, only a few publicly available datasets satisfy these criteria.
Out of all the datasets identified by the Seventh ISIC Skin Image Analysis Workshop at ECCV 2022 (derm7pt~\cite{Kawahara19}, Dermofit Image Library~\cite{ballerini2013color}, Diverse Dermatology Images (DDI)~\cite{daneshjou2022disparities}, Fitzpatrick17K~\cite{groh2021evaluating}, ISIC 2018~\cite{codella2019skin}, ISIC 2019~\cite{Tschandl18,Codella18,Combalia19}, ISIC 2020~\cite{Rotemberg21}, MED-NODE~\cite{giotis2015med}, PAD-UFES-20~\cite{pacheco2020pad}, PH2~\cite{Mendonca13}, SD-128~\cite{sun2016benchmark}, SD-198~\cite{sun2016benchmark}, SD-260~\cite{yang2022sd260}), only three datasets contain Fitzpatrick skin type labels: Fitzpatrick17K with 16,577, DDI with 656, and PAD-UFES-20 with 2,298 clinical images.
The Fitzpatrick17K dataset is the only dataset out of these three which covers all the 6 different skin types (with over 600 images per skin type) and contains more than 10K images, suitable for training high-capacity DL-based networks and our GAN-based color transformer.
It also contains samples from 114 different skin conditions, which is the largest number compared to the other two.
For these reasons, in this work, we used the Fitzpatrick17K dataset for training and evaluating our proposed method.
However, skin conditions in the Fitzpatrick17K dataset images are not verified by dermotologists and skin types in this dataset are annotated by non-dermatologists.
Also, the patient images captured in the clinical settings exhibit various lighting conditions and perspectives.
During our experiments, we found many erroneous and wrongly labeled images in the Fitzpatrick17K dataset, which could affect the training process.
Fig. \ref{fig:noise} shows some erroneous images in the Fitzpatrick17K dataset. 
Therefore, one possible future work can be cleaning the Fitzpatrick17K dataset and verifying its skin conditions and skin types by dermatologists.

In addition, as we can see in Section~\ref{sec:partial} and Figure~\ref{fig:partial}, the number of training images plays a significant role in the model's performance across different skin types.
Although in this paper we proposed a method for improving the skin condition classifier's fairness and generalizability, the importance of obtaining large and diverse datasets must not be neglected.
Mitigating bias in AI diagnosis tools in the algorithm stage, as we proposed, can be effective and is particularly essential for the currently developed models, however, future research at the intersection of dermatology and computer vision should have specific focus on adding more diverse and annotated images to existing databases.

\section{Conclusion}
In this work, we proposed \cic, a method based on domain invariant representation learning, for mitigating skin type bias in clinical image classification.
Using a domain-invariant representation learning approach and training a color-invariant model, \cic improved the accuracy for skin disease classification across different skin types for the Fitzpatrick17K dataset and set a new state-of-the-art performance on the classification of the 114 skin conditions. 
We also proposed a new fairness metric \ourM for assessing fairness of classification in the presence of multiple protected groups, and showed that \cic improves fairness of classification.
Additionally, we presented an extensive evaluation over multiple CNN backbones as well as experiments to analyze \cic's domain adaptation performance and the effect of varying the number of training images of different FST groups on its performance.

\paragraph{\textbf{Acknowledgements.}}
We would like to thank lab members Jeremy Kawahara and Ashish Sinha for their helpful discussions and comments on this work.
We would also like to thank the reviewers for their valuable feedback that helped in improving this work.
This project was partially funded by the Natural Sciences and Engineering Research Council of Canada (NSERC), and its computational resources were provided by NVIDIA and Compute Canada
(computecanada.ca).

 \bibliographystyle{splncs04}
 \bibliography{mybibliography}
\end{document}

%% file: tables/table5.tex
\begin{table*}[ht!]
\centering
\caption{Comparing the model capacities and computational requirements of different backbones evaluated. For all the six backbones, we report the number of parameters and the number of multiply-add operations (\textbf{MulAddOps}). All numbers are in millions (\textbf{M}). Note how the six backbones encompass several architectural families and a large range of model capacities ($\sim 2$M to $\sim 135$M parameters) and computational requirements ($\sim 72$M MulAddOps to $\sim 5136$M MulAddOps).}
\resizebox{1.0\textwidth}{!}{\setlength{\tabcolsep}{0.5em} 
\setlength{\tabcolsep}{0.5em}
{\renewcommand{\arraystretch}{1.35}
\begin{tabular}{lcccccc}
\toprule
               & \textbf{MobileNetV2} & \textbf{MobileNetV3L} & \textbf{DenseNet-121} & \textbf{ResNet-18} & \textbf{ResNet-50} & \textbf{VGG-16} \\ \midrule
\textbf{Parameters (M)} & 2.55                 & 4.53                  & 7.22                  & 11.31              & 24.03              & 135.31          \\
\textbf{MulAddOps (M)}  & 98.16                & 72.51                 & 925.45                & 592.32             & 1335.15            & 5136.16         \\ \bottomrule
\end{tabular}
}
}
\label{table:backboneparams}
\end{table*}

%% file: tables/table1.tex
\begin{table*}[ht!]
\centering
\caption{
Classification performance and fairness of \cic for classifying 114 skin conditions across skin types as assessed by the mean (std. dev.) of the metrics described in Section \ref{subsec:metrics}.
We compute the overall accuracy based on the micro average accuracy across all skin types. 
Values in bold indicate the best results.
\cic yields the best performance while also improving fairness.
}
\resizebox{\textwidth}{!}{%
\setlength{\tabcolsep}{0.5em} 
{\renewcommand{\arraystretch}{1.25}
\begin{tabular}{@{}lccccccccccc@{}}
\toprule
\multirow{2}{*}{\textbf{Model}}      & \multirow{2}{*}{\textbf{Recall}} & \multirow{2}{*}{\textbf{F1-score}} & \multicolumn{7}{c}{\textbf{Accuracy}}       & \multirow{2}{*}{\textbf{EOD $\downarrow$}}
& \multirow{2}{*}{\textbf{\ourMshort $\downarrow$}} 
\\ \cmidrule{4-10}                                           
& & & \textbf{Overall}                                                & \textbf{Type 1}                                                 & \textbf{Type 2}                                                 & \textbf{Type 3}                                                 & \textbf{Type 4}                                                 & \textbf{Type 5}                                                 & \textbf{Type 6}
& & 
\\ \midrule
Baseline              & 0.251 & 0.193                                             & 0.202                                                          & 0.158                                                          & 0.169                                                          & 0.222                                                          & 0.241                                                          & 0.289                                                          & 0.155                                                          
& 0.309
& 0.652                                                          \\
\hdashline
\begin{tabular}[c]{@{}l@{}}Improved\\ Baseline (Ours)\end{tabular}  &
\begin{tabular}[c]{@{}c@{}}0.444\\ (0.007)\end{tabular} &
\begin{tabular}[c]{@{}c@{}}0.441\\ (0.009)\end{tabular} &
\begin{tabular}[c]{@{}c@{}}0.471\\ (0.004)\end{tabular} & \begin{tabular}[c]{@{}c@{}}0.358\\ (0.026)\end{tabular} & \begin{tabular}[c]{@{}c@{}}0.408\\ (0.014)\end{tabular} & \begin{tabular}[c]{@{}c@{}}0.506\\ (0.023)\end{tabular} & \begin{tabular}[c]{@{}c@{}}0.572\\ (0.022)\end{tabular} & \begin{tabular}[c]{@{}c@{}}0.604\\ (0.029)\end{tabular} & \begin{tabular}[c]{@{}c@{}}0.507\\ (0.027)\end{tabular} &
\begin{tabular}[c]{@{}c@{}}0.261\\ (0.028)\end{tabular} & 
\begin{tabular}[c]{@{}c@{}}0.512\\ (0.078)\end{tabular}\\

\hdashline
\begin{tabular}[c]{@{}l@{}}\cic\\ (Ours)\end{tabular}  &
\begin{tabular}[c]{@{}c@{}}\textbf{0.459}\\ (0.003)\end{tabular} & \begin{tabular}[c]{@{}c@{}}\textbf{0.459}\\ (0.003)\end{tabular} &
\begin{tabular}[c]{@{}c@{}}\textbf{0.488}\\ (0.005)\end{tabular} & \begin{tabular}[c]{@{}c@{}}\textbf{0.379}\\ (0.019)\end{tabular} & \begin{tabular}[c]{@{}c@{}}\textbf{0.423}\\ (0.011)\end{tabular} & \begin{tabular}[c]{@{}c@{}}\textbf{0.528}\\ (0.024)\end{tabular} & \begin{tabular}[c]{@{}c@{}}\textbf{0.592}\\ (0.022)\end{tabular} & \begin{tabular}[c]{@{}c@{}}\textbf{0.617}\\ (0.021)\end{tabular} & \begin{tabular}[c]{@{}c@{}}\textbf{0.512}\\ (0.043)\end{tabular} & 
\begin{tabular}[c]{@{}c@{}}\textbf{0.252}\\ (0.031)\end{tabular} & 
\begin{tabular}[c]{@{}c@{}}\textbf{0.474}\\ (0.047)\end{tabular}\\ \bottomrule
\end{tabular}%
}
}
\label{table:result}
\end{table*}

%% file: tables/table2.tex
\begin{table*}[ht!]
\centering
\caption{Evaluating the classification performance improvement contribution of the regularization loss $\mathcal{L}_{reg}$ with multiple different feature extractor backbones. Reported values are the mean (std. dev.) of the metrics described in Section \ref{subsec:metrics}. Best values for each backbone are presented in bold. 
EOD reported (for two groups of light and dark FSTs) for completeness but evaluation over all the 6 FSTs uses \ourMshort (see text for details). 
Observe that $\mathcal{L}_{reg}$ improves the classification accuracy and the fairness metric \ourMshort for all backbones.}
\resizebox{\textwidth}{!}{%
\setlength{\tabcolsep}{0.5em} 
{\renewcommand{\arraystretch}{1.35}
\begin{tabular}{@{}lcccccccccccc@{}}
\toprule
\multirow{2}{*}{\textbf{Model}} & \multirow{2}{*}{\textbf{$\mathcal{L}_{reg}$}}      & \multirow{2}{*}{\textbf{Recall}} & \multirow{2}{*}{\textbf{F1-score}} & \multicolumn{7}{c}{\textbf{Accuracy}}       & \multirow{2}{*}{\textbf{EOD $\downarrow$}}
& \multirow{2}{*}{\textbf{\ourMshort $\downarrow$}} 
\\ \cline{5-11}                                           
& & & & \textbf{Overall}                                                & \textbf{Type 1}                                                 & \textbf{Type 2}                                                 & \textbf{Type 3}                                                 & \textbf{Type 4}                                                 & \textbf{Type 5}                                                 & \textbf{Type 6}
& & 

\\ \midrule
\multirow{2}{*}{MobileNetV2}  & \xmark  & 0.375 & 0.365        & 0.398           & 0.313          & \textbf{0.364}          & 0.409          & 0.503          & 0.491          & 0.333          
& 0.280
& 0.472       \\
                              & \cmark  & \textbf{0.404} & \textbf{0.397}             & \textbf{0.434}           & \textbf{0.354}          & 0.357          & \textbf{0.471}          & \textbf{0.559}          & \textbf{0.544}          & \textbf{0.421}          
                              & 0.258
                              & \textbf{0.455}       \\ \hdashline 
\multirow{2}{*}{MobileNetV3L} & \xmark   & \textbf{0.427} & 0.403           & 0.438           & 0.357          & 0.388          & 0.449          & 0.543          & \textbf{0.560}          & 0.413          
& 0.271
& 0.449       \\
                              & \cmark & 0.425 & \textbf{0.412}              & \textbf{0.451}           & \textbf{0.369}          & \textbf{0.400}          & \textbf{0.464}          & \textbf{0.565}          & 0.550          & \textbf{0.444}          
                              & 0.275
                              & \textbf{0.420}       \\ \hdashline 
\multirow{2}{*}{DenseNet-121} & \xmark & 0.425 & 0.416             & 0.451           & 0.393          & 0.397          & 0.452          & \textbf{0.565}          & 0.522          & \textbf{0.500}          
& 0.278
&  0.364       \\
                              & \cmark & \textbf{0.441} & \textbf{0.430}              & \textbf{0.462}           & \textbf{0.413}          & \textbf{0.406}          & \textbf{0.473}          & 0.561          & \textbf{0.550}          & 0.452          
                              & 0.294
                              &  \textbf{0.324}       \\ \hdashline 
\multirow{2}{*}{ResNet-18}    & \xmark & 0.391 & 0.381               & 0.417           & 0.355          & 0.353          & 0.431          & 0.538          & 0.516          & \textbf{0.389}          
& 0.263
& 0.430       \\
                              & \cmark & \textbf{0.416} & \textbf{0.410}               & \textbf{0.436}           & \textbf{0.367}          & \textbf{0.380}          & \textbf{0.458}          & \textbf{0.543}          & \textbf{0.538}          & \textbf{0.389}          
                              & 0.282
                              & \textbf{0.395}       \\ \hdashline 
\multirow{2}{*}{ResNet-50}    & \xmark & 0.390 & 0.382              & 0.416           & 0.337          & 0.363          & 0.422          & 0.549          & 0.506          & 0.389          
& 0.257
& 0.497       \\
                              & \cmark & \textbf{0.440} & \textbf{0.429}              & \textbf{0.466}           & \textbf{0.384}          & \textbf{0.402}          & \textbf{0.502}          & \textbf{0.580}          & \textbf{0.569}          & \textbf{0.421}          
                              & 0.283
                              & \textbf{0.411}       \\  
                              \bottomrule
\end{tabular}%
}
}
\label{table:backbones}
\end{table*}

%% file: tables/table3.tex
\begin{table*}[h!]
\centering
\caption{Classification performance measured by micro average accuracy when trained and evaluated on holdout sets composed of different Fitzpatrick skin types (FSTs). For example, ``FST3-6'' denotes that the model was trained on images only from FSTs 1 and 2 and evaluated on FSTs 3, 4, 5, and 6. \cic achieves higher classification accuracies than Baseline (Groh et al.~\cite{groh2021evaluating}) and Improved Baseline (also ours) for all holdout partitions and for all skin types.}
\resizebox{\textwidth}{!}{%
\setlength{\tabcolsep}{0.5em} 
{\renewcommand{\arraystretch}{1.35}
\begin{tabular}{@{}ccccccccc@{}}
\toprule
\textbf{\begin{tabular}[c]{@{}c@{}}Holdout\\ Partition\end{tabular}} & \textbf{Method}                      & \textbf{Overall} & \textbf{Type 1} & \textbf{Type 2} & \textbf{Type 3} & \textbf{Type 4} & \textbf{Type 5} & \textbf{Type 6} \\ \midrule
\multirow{3}{*}{FST3-6}                                                 & Baseline & 0.138            & -               & -               & 0.159           & 0.142           & 0.101           & 0.090           \\
                                                                     & Improved Baseline                                  & 0.249           & -               & -               & 0.308           & 0.246           & 0.185           & 0.113           \\
                                                                     & \cic                                  & \textbf{0.260}            & -               & -               & \textbf{0.327}           & \textbf{0.250}           & \textbf{0.193}           & \textbf{0.115}           \\   \hdashline
\multirow{3}{*}{FST12 and FST56}                                                 & Baseline & 0.134            & 0.100           & 0.130           & -               & -               & 0.211           & 0.121           \\
                                                            & Improved Baseline                                  & 0.272            & 0.181           & 0.274           & -               & -               & 0.453           & 0.227           \\
                                                                     & \cic                                  & \textbf{0.285}            & \textbf{0.199}           & \textbf{0.285}           & -               & -               & \textbf{0.469}           & \textbf{0.233}           \\   \hdashline
\multirow{3}{*}{FST1-4}                                                 & Baseline & 0.077            & 0.044           & 0.055           & 0.091           & 0.129           & -               & -               \\
                                                            & Improved Baseline & 0.152            & 0.078           & 0.111           & 0.167           & 0.280           & -               & -               \\ 
                                                                     & \cic                                  & \textbf{0.163}            & \textbf{0.095}           & \textbf{0.121}           & \textbf{0.177}           & \textbf{0.293}           & -               & -               \\ \bottomrule
\end{tabular}
}
}
\label{table:2skin}
\end{table*}

%% file: tables/table4.tex
\begin{table*}[h!]
\centering
\caption{Total number of training images for each experiment illustrated in Figure~\ref{fig:partial}. Note that the test set for all these experiments is the original test split with 3,205 images ($20\%$ of the Fitzpatrick17K dataset images), and the number of training images for experiments with $100\%$ of each FST group is the same for all three groups, and is equal to the original train split with 11,934 images ($70\%$ of the Fitzpatrick17K dataset images).}
\setlength{\tabcolsep}{1.0em}
{\renewcommand{\arraystretch}{1.2}
\begin{tabular}{lccccc}
\toprule
      & \textbf{0\%}     & \textbf{20\%}     & \textbf{40\%}     & \textbf{60\%}     & \textbf{80\%}     \\ \midrule
\textbf{FST12} & 5,964 & 7,073  & 8,183  & 9,293  & 10,403 \\
\textbf{FST34} & 7,088 & 7,973  & 8,858  & 9,743  & 10,628 \\
\textbf{FST56} & 9,974 & 1,0281 & 10,589 & 10,897 & 11,205 \\ \bottomrule
\end{tabular}
}
\label{table:partial}
\end{table*}